\UseRawInputEncoding
\pdfoutput=1

\documentclass[11pt]{article}

\usepackage[dvipsnames, svgnames, x11names]{xcolor}
\usepackage{acl}

\usepackage{times}
\usepackage{latexsym}

\usepackage[T1]{fontenc}

\usepackage[utf8]{inputenc}

\usepackage{microtype}

\usepackage{booktabs}

\usepackage{amsmath}
\usepackage{url}
\usepackage{multirow}
\usepackage{graphicx}
\usepackage{color}

\title{CHAE: Fine-Grained Controllable Story Generation with Characters, Actions and Emotions}

\author{Xinpeng Wang\textsuperscript{1}, Han Jiang\textsuperscript{1}, Zhihua Wei\textsuperscript{1}\thanks{~~Corresponding author}, Shanlin Zhou\textsuperscript{2} \\ 
 \textsuperscript{1}Department of Computer Science and Technology, Tongji University, Shanghai, China \\ 
 \textsuperscript{2}School of Computer Science and Technology, Shanghai University of Electric Power, Shanghai, China \\
  \texttt{\{wangxinpeng, 2230780, zhihua\_wei\}@tongji.edu.cn} \\
  \texttt{zhoushanlin@mail.shiep.edu.cn} }

\begin{document}
\maketitle
\begin{abstract}
Story generation has emerged as an interesting yet challenging NLP task in recent years. Some existing studies aim at generating fluent and coherent stories from keywords and outlines; while others attempt to control the global features of the story, such as emotion, style and topic. However, these works focus on coarse-grained control on the story, neglecting control on the details of the story, which is also crucial for the task. To fill the gap, this paper proposes a model for fine-grained control on the story, which allows the generation of customized stories with characters, corresponding actions and emotions arbitrarily assigned. Extensive experimental results on both automatic and human manual evaluations show the superiority of our method. It has strong controllability to generate stories according to the fine-grained personalized guidance, unveiling the effectiveness of our methodology. Our code is available at \url{https://github.com/victorup/CHAE}.


\end{abstract}

\section{Introduction}

Story generation, one of emergent tasks in the field of natural language generation, requires following sentences given the beginning of the story. For human beings, it is believed that storytelling requires strong logical thinking ability and organizational competence, and for machines it may be even more intractable. Nonetheless, works on story generation can help machines communicate with humans and drive improvements in natural language processing \citep{alabdulkarim-etal-2021-automatic}.

At present, most works on story generation focus on the coherence of the story generated according to keywords, outlines and commonsense knowledge \citep{yao2019plan, guan2019story, rashkin-etal-2020-plotmachines, guan2020knowledge, ji2020language}. Some other works aim at generating stories controlled by overall emotion, style, and topic \citep{keskar2019ctrl, xu2020controllable, brahman-chaturvedi-2020-modeling, kong-etal-2021-stylized}. However, in reality, people often expect more detailed designs catering to their needs rather than a simple theme or topic in the generated story. For example, a novel with more complete elements, i.e., plot, character, theme, viewpoint, symbol, and setting is usually preferred to those made up out of thin air.

Taking the control in story generation as the cutting point, GPT-2 \citep{radford2019language} can fulfill the story according to the beginning, but the process of generation cannot be controlled by people, resulting in unlogical outputs that lack practicality.
CTRL \citep{keskar2019ctrl} can specify the generation of articles with different styles through some style words, but such control stays at the coarse-grained level, and makes a relatively weak influence. CoCon \citep{chan2020cocon} introduces natural language to guide text generation. \citet{fang2021outline} propose a new task that guides paragraph generation through a given sequence of outline events. However, the above two studies just explicitly add some contents to the generated sentences, which is similar to forming sentences with given phrases, not using the input as a condition guide for the generative models.
SoCP presented by \citet{xu2020controllable} can generate stories under changeable psychological state control, while it does not govern the detailed contents of the story.

In this paper, we consider more fine-grained control on story generation, and propose a model, \textbf{CHAE} for fine-grained controllable story generation, allowing the generation of stories with customized \textbf{CH}aracters, and their \textbf{A}ctions and \textbf{E}motions. Characters are the core of the story. Their actions drive the story along, and their emotions make the story lively and interesting. Consequently, we take the characters along with their actions and emotions as the control conditions. It is a challenge that our model needs to control multiple characters with their actions and emotions respectively in a story, especially under the guidance in the form of natural language. To crack the nut, a novel input form conducive to fine-grained control on story generation is introduced into CHAE. Concretely, we use various prompts for fine-grained control conditions in different aspects. 
Moreover, we design different methods for different control conditions to improve the control effect.
Inspired by multi-task learning, we incorporate a character-wise emotion loss while training, thus enforcing the relevance between the characters and their emotions respectively.

The contributions of our work can be summarized as follows:

\begin{itemize}
\item 
We first take the characters with their actions and emotions of the story into account to conduct more fine-grained controllable story generation.

\item 
We propose a model CHAE with a novel input form that helps the model control the story in various aspects, and a character-wise emotion loss to relate the characters and the corresponding emotions.

\item 
The results of both automatic and human evaluation show that our model has strong controllability to generate customized stories.
\end{itemize}

\section{Related Work}

\textbf{Story Generation}~~
Story generation has attracted more and more researchers to explore in recent years. 
There are many challenges in the task, such as context coherence and control. 
For context coherence, some works are devoted to introducing a series of keywords \citep{yao2019plan}, outlines \citep{rashkin-etal-2020-plotmachines}, or incorporating external knowledge \citep{guan2019story,guan2020knowledge,ji2020language} into the story.
For style and sentiment control, \citet{kong-etal-2021-stylized} generate stories with specified style given a leading context. However, it only focuses on the global attributes of the story.
\citet{brahman-chaturvedi-2020-modeling} work on generating stories with desired titles and the protagonists' emotion arcs, and \citet{xu2020controllable} generate stories considering the changes in the psychological state, while they just control the emotion lines instead of the detailed contents.

~\\
\noindent
\textbf{Controllable Text Generation}~~
We have witnessed the great performance of SOTA models for text generation these years. Despite the progress in coherence and rationality of the text generated, controllability remains to be challenging, which means generating text with specific attributes, such as emotion, style, topic, format, etc.
CTRL \citep{keskar2019ctrl} can control the overall attributes such as domain, style and topic of the generated text by adding control codes and prompts. 
By plugging in a discriminator, PPLM \citep{Dathathri2020Plug} can guide text generation without further training the language model. 
CoCon \citep{chan2020cocon} fine-tunes an intermediate block with self-supervised learning to control high-level attributes i.e., sentiment and topic. 
Compared to the previous works, our work places the emphasis on more fine-grained, all-round control on the generating process, including the control of characters with their emotions and actions in the story.

\begin{figure}[t]
\centering
\includegraphics[scale=0.43]{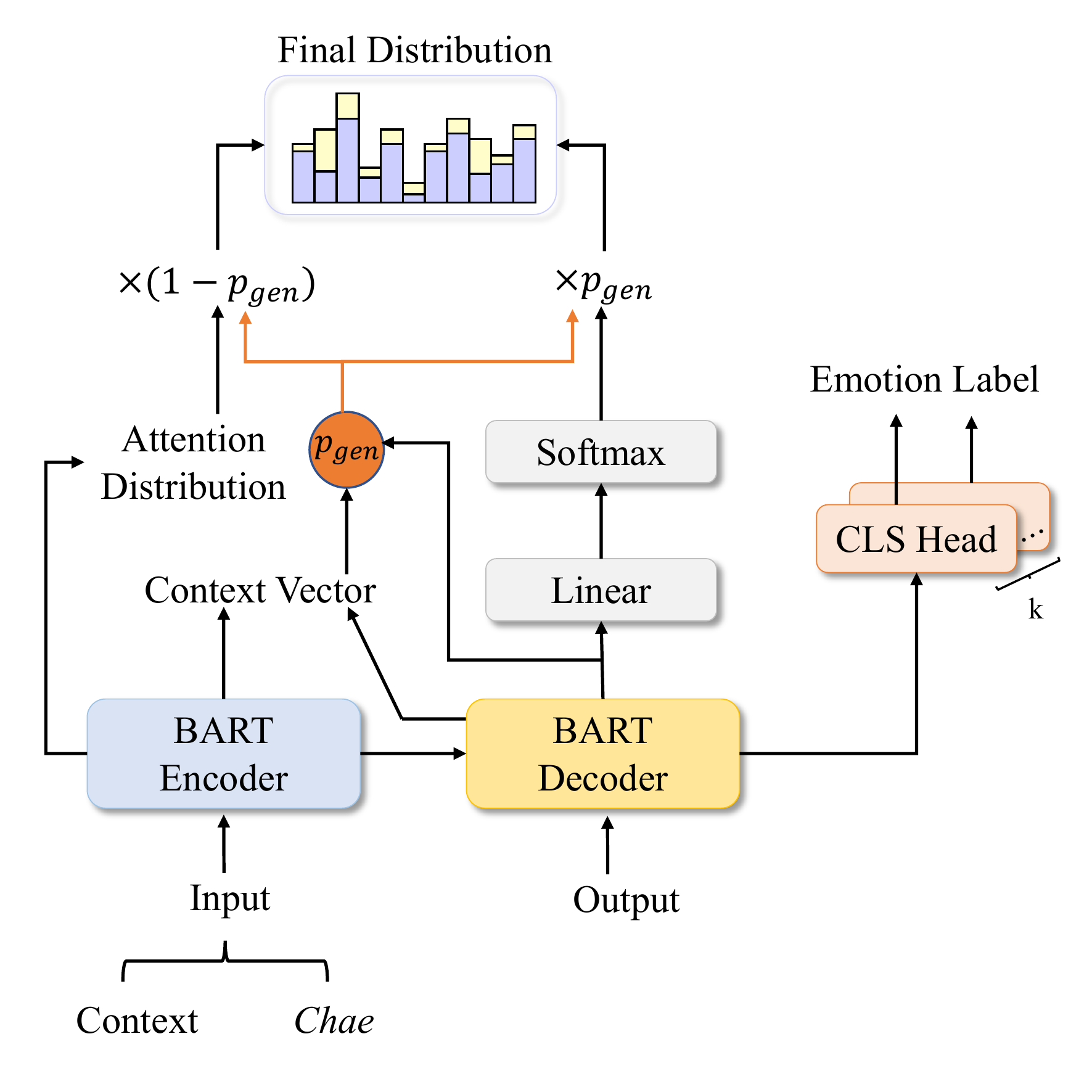}
\caption{The architecture of CHAE. The input is the concatenation of two components, $Context$ and $Chae$, which will be further explained in Sec \ref{sec3}. The emotion labels are used for calculating a character-wise emotion loss to tie up the characters and their emotions respectively.} 
\label{model_pic}
\end{figure}

\begin{figure*}[t]
\centering
\includegraphics[scale=0.45]{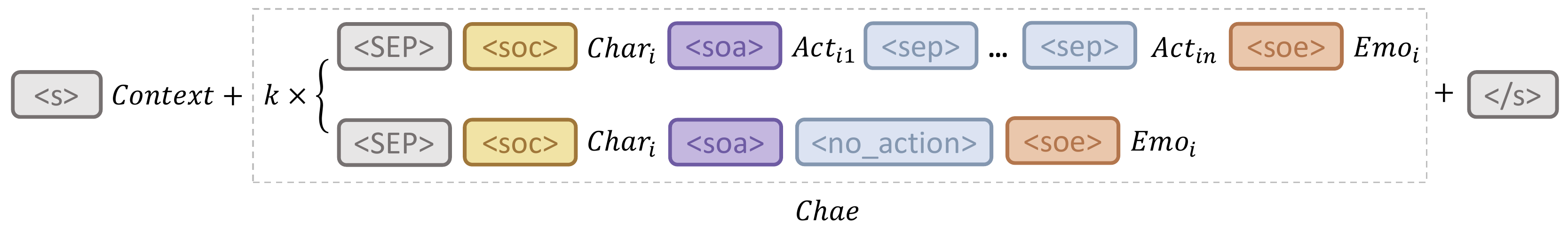}
\caption{The input form of CHAE. The input starts with $\langle s \rangle$ and ends with $\langle /s \rangle$, comprising $Context$ and $Chae$. The latter is a sequence of $k$ control conditions on the next sentence to be generated, and each condition controls a character. We show two possible forms of the control conditions after the brace. The special tokens in $Chae$ are further explained in Table \ref{Chae explanation}.}
\label{input_chae}
\end{figure*}

\section{Methodology}
\label{sec3}

\subsection{Problem Formulation}
The process of fine-grained controllable story generation in this work is defined as follows.

The input of the task has two components. We refer to the one as $Context$. Let $Context = (x{_1},x{_2},...,x{_p})$ denote the beginning sentence of the story, which will be the initial $Context$. The other component is $Chae$, a sequence of $k$ fine-grained control conditions on the next sentence. Each condition in $Chae$ is the combination of the name $Char_i$, $n$ actions $Act_{i1}, Act_{i2}, ..., Act_{in}$, and emotion $Emo_i$ of a character appearing in the next sentence to be generated, where $n$ is not fixed and $i$ is the index of the character. Note that we use Italic $Chae$ to distinguish the special input component from our model \textbf{CHAE}.

The model predicts one sentence denoted as $Y = (y{_1},y{_2},...,y{_q})$ at a time by estimating the conditional probability $P(Y|Context, Chae)$. Here we embody the idea of auto-regression by adopting an iterative generation strategy, that is, the sentence generated is then concatenated to $Context$ for next prediction. Especially, at training time, we concatenate the gold sentences instead of generated sentences to $Context$ incrementally like teacher forcing.

The goal of this task is to generate a story where each sentence adheres to the input condition $Chae$ in terms of character, action, and emotion, through which elevate the quality of generation in a fine-grained manner.

\subsection{Model Architecture}
The architecture of our model CHAE is shown in Figure \ref{model_pic}.
CHAE is built upon a BART model \citep{lewis-etal-2020-bart}. As mentioned, our model embodies the idea of auto-regression by adopting an iterative generation strategy. On the one hand, the iteratively updated $Chae$ helps control the content of each sentence at a detailed level of granularity. On the other hand, the strategy ensures that the model can always see the foregoing. Considering the incremental $Context$ can be extra long, we employ BART rather than GPT-2. GPT-2 is an auto-regressive model fully based on transformer decoder, while BART has a bidirectional encoder, which might make it better in understanding and encoding long input sequences. To confirm the hypothesis, we also compared BART with GPT-2 on the benchmark dataset in Sec \ref{sec4}, and found that BART outperformed GPT-2 in story generation.

\begin{table}[t]
\centering
\resizebox{\linewidth}{!}{
\begin{tabular}{ll}  
\toprule
\textbf{Special tokens} & \textbf{Meaning}\\
\midrule
$\langle SEP \rangle$ & The start token of a condition. \\
$\langle soc \rangle$ & The start token of a character's name.\\
$\langle soa \rangle$ & The start token of actions.\\
$\langle soe \rangle$ & The start token of an emotion.\\
$\langle sep \rangle$ & The start token of a single action.\\
$\langle no\_action \rangle$ & The token representing no action.\\
\bottomrule
\end{tabular}
}
\caption{\label{Chae explanation}
The meanings of the special tokens in $Chae$.
}
\end{table}

\subsection{Generation Based on Fine-Grained Control Conditions}
To generate a story with the characters, their actions and emotions specified, we need to remind the BART model of the elements controlled currently from time to time. Inspired by the practice of leveraging special tokens for controllable generation \citep{fang2021outline, keskar2019ctrl, DBLP:journals/corr/TsutsuiC17a}, we propose a novel form of input (titled $Chae$), which is a sequence of $k$ fine-grained control conditions on the next sentence to be generated. Each condition in $Chae$ controls a character, and each segment in the condition controls an element (i.e., character's name, action, and emotion) of the corresponding character. Note that any number of actions can be assigned in a condition. The nested sequence form of $Chae$ facilitates the neat combination of various fine-grained control conditions.

We design several special tokens and add them between each segment as the control prompts (see Figure \ref{input_chae}). 
In this study, 6 special tokens are used to prompt the model. They are
$\langle SEP \rangle$, $\langle soc \rangle$, $\langle soa \rangle$, $\langle soe \rangle$, $\langle sep \rangle$, and $\langle no\_action \rangle$. The meanings of the tokens are shown in Table \ref{Chae explanation}.

Then, we encode the input $Context$ and $Chae$ as $h_{enc}$ by the BART encoder: 
\begin{equation}
\boldsymbol{h}_{enc} = \mathbf{Enc}(\boldsymbol{e}_c),
\end{equation}
where $\boldsymbol{e}_c$ is the embedding of $Context$ and $Chae$. The vocab generation probability is calculated by the BART decoder as:
\begin{equation}
\begin{aligned}
P_{voc}(y) &= P(y_t|y_{<t}, Context, Chae) \\&= \mathrm{softmax}(\boldsymbol{W}_{voc}\boldsymbol{h}_{dec}),
\end{aligned}
\end{equation}
\begin{equation}
\boldsymbol{h}_{dec} = \mathbf{Dec}(\boldsymbol{e}_{y<t}, \boldsymbol{h}_{enc}),
\end{equation}
where $\boldsymbol{e}_{y<t}$ is the embeddings of the generated tokens before timestep $t$, and $\boldsymbol{W}_{voc}$ is a trainable parameter.

After fine-tuning with these special tokens, the model is aware of the elements controlled by each segment of each condition.

\subsection{Improvement in Control Effect}
\subsubsection{Character and Action Control}
For characters and their actions, we expect to see the characters and their actions in current $Chae$ appear in the coming sentence. Inspired by the usage of copy mechanism \citep{see-etal-2017-get, deaton2019transformers, prabhu2020making} in copying significant tokens from the input sequences, we add a copy pointer to BART for the information in $Chae$. 
The attention distribution on $Chae$ denoted by $\widetilde{a}$ is attained by averaging the multiple heads in the cross attention block of BART decoder:
\begin{equation}
\widetilde{a} = \frac{\sum_{i=1}^{h}{a_i}}{h},
\end{equation}
where $h$ is the number of the attention heads, and $a_i$ denotes the attention distribution on $Chae$ from the $i$-$th$ attention head. 

When generating stories, we first combine the hidden state of the decoder $\boldsymbol{h}_{dec}$, the context vector $\boldsymbol{h}_{con}$, and the embedding of the decoder input $\boldsymbol{e}_y$. Secondly, we calculate a generation probability $p_{gen}$, which is a soft switch to choose a word from the vocabulary according to $P_{voc}$, or to copy a word from $Chae$ by sampling from the mean attention distribution $\widetilde{a}$. The final distribution of a word can be represented as follows:
\begin{equation}
P(y)=p_{gen} P_{voc}(y)+\left(1-p_{gen}\right) \sum_{j: y_j=y}\widetilde{a_{j}},
\end{equation}
\begin{equation}
p_{g e n}=\sigma(\boldsymbol{W}_{p}^{\top}[\boldsymbol{h}_{dec}; \boldsymbol{h}_{con}; \boldsymbol{e}_{y}]),
\end{equation}
where $\boldsymbol{W}_{p}$ is a trainable parameter. 
In this way, the characters and actions in $Chae$ will be produced with higher probability, and the output can be flexibly changed according to the different input.

\begin{figure}[t]
\centering
\includegraphics[scale=0.3]{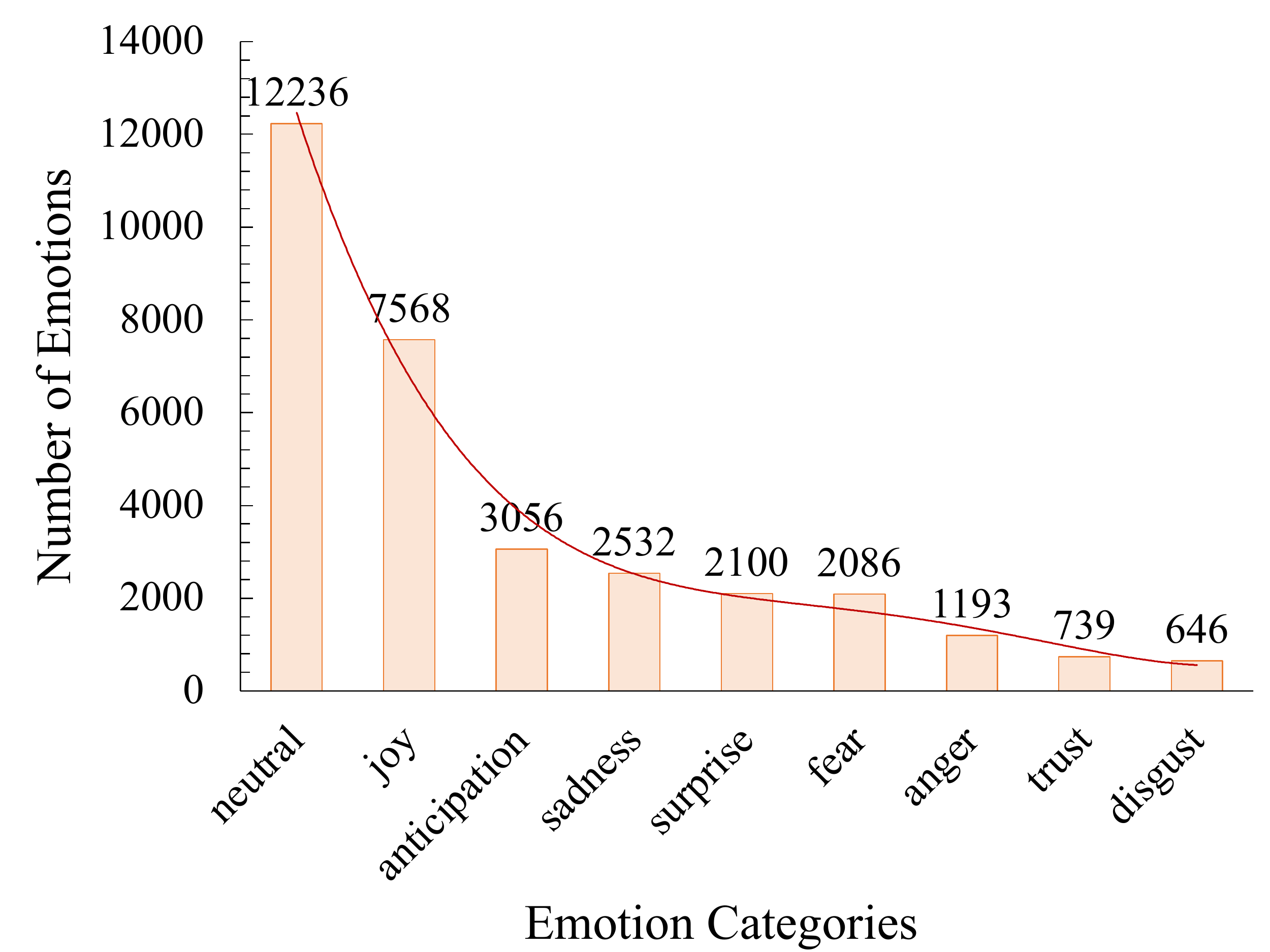}
\caption{Statistics of emotion categories of sentences in all stories.} 
\label{emotions_statistic}
\end{figure}

\subsubsection{Character-Wise Emotion Control}
For characters' emotions, we additionally incorporate a character-wise emotion loss, by which the model is forced to generate sentences with specified emotions tied up with corresponding characters. We add $k$ emotion classification heads to the top of the decoder output layer \citep{ide-kawahara-2021-multi}, and $k$ is equivalent to the number of conditions in $Chae$. It provides direct supervision on emotion control to predict the emotion of every character in the story.

However, as shown in Figure \ref{emotions_statistic}, the emotion categories present a long tail distribution. To relieve the class-imbalance problem, we use a Weighted Cross-Entropy (WCE) loss between the predicted emotion distribution $P_{emo}$ and the emotion labels $l_e$:
\begin{equation}
L_{EMO}=- \alpha_e l_e \log (P_{emo}),
\end{equation}
\begin{equation}
P_{emo} = \mathrm{softmax}(\boldsymbol{W}_{emo} \boldsymbol{h}_{dec}),
\end{equation}
\begin{equation}
\alpha_e = \frac{N}{e * \mathrm{count}(l_{e})},
\end{equation}
where $\boldsymbol{W}_{emo}$ is a trainable parameter, $\alpha_e$ denotes the weights of emotion classification labels. $N$ is the number of the training samples, $e$ is the number of emotion categories, and count(·) is a function to calculate the number of samples in each emotion category. Now, the model has both explicit control from emotion in $Chae$ and implicit control from the emotion loss to generate sentences with target emotions.

\subsection{Objective Function}
We minimize the Negative Log-Likelihood (NLL) loss of the target story sentence $Y$ with the input $Context$ and $Chae$:
\begin{equation}
L_{NLL}=-\sum_{t=1}^{T} \log P(y_{t}|y_{<t},Context,Chae),
\end{equation}
The total loss $L$ is as follow:
\begin{equation}
L = L_{NLL} + \lambda L_{EMO},
\end{equation}
where $\lambda$ is a hyper-parameter. After training with the above two objectives, our model can generate fluent stories under desired conditions.

\section{Experiments}
\label{sec4}

\subsection{Dataset}
We use ROCStory with labeled characters' emotions and actions as our dataset \citep{rashkin-etal-2018-modeling}, which contains 14738 five-sentence stories, including 9885 stories for training, 2483 stories for validation and 2370 stories for testing.
To conform to the iterative generation, we divide the stories into 39540 / 9932 / 9480 sentence pairs for training/validation/testing. Each pair of sentences consists of two adjacent sentences in the story.
The characters with their actions and emotions in the dataset are labeled by three crowdsourced workers from Amazon Mechanical Turk. The emotions come from Plutchik psychology theory \citep{plutchik1980general}, including 8 species, such as “joy”, “anger”, etc. Most actions represent the character's underlying motivations, and they generally take the infinitive such as “to win all games” and “to have fun”.

In preprocessing, we integrate the annotations of the three workers, and take the emotions with the highest confidence as the final labels. However, we still notice that some emotion labels have excessively low confidence, indicating subtle emotion tendencies. We modify them to “neutral” to avoid distortion of the emotions.
Moreover, we find that 9885 stories in the training set are not labeled with emotions, but only with actions. As a remedy, we fine-tune the model on these 9885 stories to keep the fluency of the story. Later, we re-divide all the stories with labeled emotions in the validation set and testing set into another 3 splits. Finally, we fine-tune the model on the new training set to capture emotional information.
The re-divided dataset includes 15528 / 1944 / 1940 sentence pairs by 8:1:1 for training/validation/testing. 

\begin{figure}[t]
\centering
\includegraphics[scale=0.3]{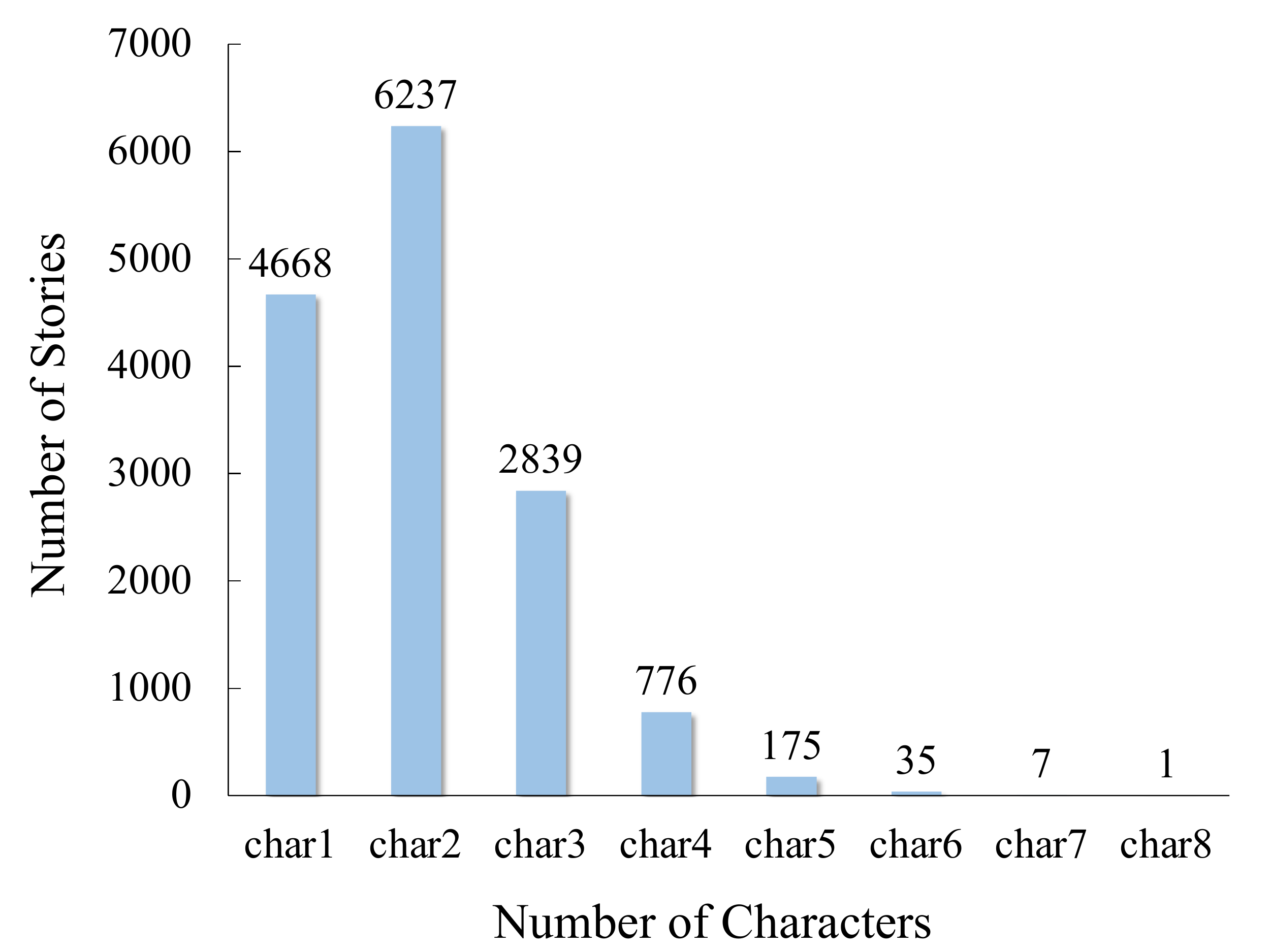}
\caption{Statistics of stories with different numbers of characters. The “char1” means that the stories have one character.} 
\label{characters_statistic}
\end{figure}

\subsection{Baselines}

We compare our model with a carefully selected set of baselines as shown below.

\textbf{GPT-2} \citep{radford2019language}:
GPT-2 is a transformer-based model pre-trained on a very large corpus, which is very commonly used in natural language generation. 
A lot of works witness a good performance of GPT-2 in dialogue, story and other text generation in recent years, demonstrating its auto-regression quality.

\textbf{BART} \citep{lewis-etal-2020-bart}:
BART is a transformer-based seq2seq pre-training model with a bidirectional (BERT-like) encoder and an autoregressive (GPT-like) decoder. 
A lot of text generation tasks, like neural machine translation and automatic summarization, can achieve effective results by fine-tuning on BART.

\textbf{SoCP} \citep{xu2020controllable}:
Stories with multi-characters and multi-psychology generated by SoCP can change with the emotional lines of assigned characters. In addition, SoCP can generate stories with different emotional intensities. It also designe a metric to evaluate the accuracy of controlling emotions of roles.

\textbf{Stylized-Story-Generation (SSG)} \citep{kong-etal-2021-stylized}:
SSG can generate stories with specified style given a leading context by first planning the stylized keywords and then generating the whole story with the guidance of the keywords. 
Two story styles are considered in SSG, including emotion-driven and event-driven stories.


\subsection{Implementation Details}

We build our model based on BART using the Huggingface's Transformers  library in Pytorch \citep{wolf2019huggingface}. We initialize our model with the public checkpoint of bart-large-cnn \footnote{\url{https://huggingface.co/facebook/bart-large-cnn/tree/main}}. The batch size during training is 8. We use the AdamW optimization \citep{DBLP:conf/iclr/LoshchilovH19} with $\beta_1 = 0.9$, $\beta_2 = 0.999$ and the initial learning rate is $5e-5$. According to the statistics of characters in the stories shown in Figure \ref{characters_statistic}, most stories contain two characters, so we fix the number of characters in each story $k$ to $2$. The hyper-parameter $\lambda$ defaults to $1.0$.

For all models, We generate stories by using top-k sampling \citep{fan-etal-2018-hierarchical} with $k = 50$ and a softmax temperature of $0.8$.

\begin{table}[t]
\centering
\resizebox{\linewidth}{!}{
\begin{tabular}{ccccccc}  
\toprule
\textbf{Models} & \textbf{PPL ↓} & \textbf{B-1 ↑} & \textbf{B-2 ↑} & \textbf{D-1 ↑} & \textbf{D-2 ↑}  &  \textbf{ACC ↑}\\
\midrule
SoCP & 101.33 & 22.93 & 7.32 & 0.478 & 0.650 & 0.893 \\
SSG & 15.22 & 26.74 & 10.77 & 0.573 & 0.904 & - \\
GPT-2 & 29.21 & 21.30 & 6.35 & 0.744 & 0.960 & - \\
BART & 14.00 & 24.15 & 7.93 & 0.729 & 0.964 & - \\
\midrule
CHAE & \textbf{11.58} & 27.10 & 10.20 & 0.750 & 0.971 & \textbf{0.941} \\
w/o copy & 11.65 & 26.53 & 9.67 & \textbf{0.754} & \textbf{0.972} & 0.879 \\
w/o emo & 11.75 & \textbf{27.51} & 10.61 & 0.732 & 0.965 & - \\
w/o copy w/o emo & 11.70 & 27.19 & \textbf{10.79} & 0.735 & 0.967 & - \\
\bottomrule
\end{tabular}}
\caption{\label{auto-metrics}
Automatic metrics.
}
\end{table}

\begin{table}[t]
\centering
\resizebox{\linewidth}{!}{
\begin{tabular}{ccccc}  
\toprule
\textbf{Setting} & \textbf{B-1 ↑} & \textbf{B-2 ↑} & \textbf{D-1 ↑} & \textbf{D-2 ↑}\\
\midrule
Greedy & 31.08 & 15.65 & 0.586 & 0.797 \\
Beam=2 & 31.96 & 15.61 & 0.595 & 0.811 \\
Beam=3 & \textbf{32.04} & \textbf{15.89} & 0.584 & 0.804 \\
Beam=4 & 31.86 & 15.84 & 0.578 & 0.801 \\
Beam=5 & 31.60 & 15.40 & 0.576 & 0.802 \\
Top-k=30, Temperature=0.8 & 30.86 & 13.63 & 0.697 & 0.939 \\
Top-k=50, Temperature=0.8 & 30.58 & 13.64 & 0.702 & 0.939 \\
Top-k=30, Temperature=1 & 29.36 & 12.00 & 0.720 & 0.957 \\
Top-k=50, Temperature=1 & 29.22 & 11.94 & 0.730 & 0.962 \\
Top-k=30, Temperature=1.2 & 27.68 & 10.49 & 0.745 & 0.968 \\ 
Top-k=50, Temperature=1.2 & 27.10 & 10.20 & \textbf{0.750} & \textbf{0.971} \\
\bottomrule
\end{tabular}}
\caption{\label{Decoding-strategies}
Decoding strategies adjustment.
}
\end{table}

\begin{table*}[t]
\centering
\resizebox{\linewidth}{!}{
\begin{tabular}{c|ccc|ccc|ccc}  
\toprule
\multirow{2}{*}{\textbf{Models}} & \multicolumn{3}{c|}{\textbf{Fluency}} & \multicolumn{3}{c|}{\textbf{Coherence}} & \multicolumn{3}{c}{\textbf{Informativeness}}\\
& \textbf{Win}(\%) & \textbf{Lose}(\%) & \textbf{Tie}(\%) & \textbf{Win}(\%) & \textbf{Lose}(\%) & \textbf{Tie}(\%) & \textbf{Win}(\%) & \textbf{Lose}(\%) & \textbf{Tie}(\%) \\
\midrule
CHAE vs. SSG & \textbf{56.0} & 30.0 & 14.0  & \textbf{52.0} & 30.0 & 18.0  & \textbf{90.0} & 6.0 & 4.0 \\
CHAE vs. BART & \textbf{40.0} & 40.0 & 20.0  & \textbf{38.0} & 36.0 & 26.0  & \textbf{46.0} & 42.0 & 12.0 \\
CHAE vs. GPT2 & \textbf{38.0} & 36.0 & 26.0  & \textbf{40.0} & 40.0 & 20.0  & \textbf{72.0} & 12.0 & 16.0 \\
\midrule
\textbf{Controllability of CHAE} & \textbf{65.0\%}
\\
\bottomrule
\end{tabular}
}
\caption{\label{Human}
Human evaluation results. It shows the percentage of win, lose and tie of CHAE compared with other baselines. The controllability of CHAE indicates the average percentage of sentences in a story that can be controlled by control condition $Chae$.
}
\end{table*}

\subsection{Automatic Evaluation}

\textbf{Evaluation Metrics}~~
We use the following metrics for automatic evaluation: 
(1) \textbf{Perplexity (PPL)}: PPL represents the general quality of the generated stories, which estimates the probability of sentences according to each word.
(2) \textbf{BLEU (B-n)} \citep{papineni-etal-2002-bleu}: We use BLEU to compare the coverage n-gram in the candidate stories and the reference stories because the words in $Chae$ usually appear in the reference stories.
(3) \textbf{Distinct (D-n)}: \citep{li-etal-2016-diversity}: Distinct is used to evaluate generation diversity by calculating the percentage of unique n-grams.
(4) \textbf{Accuracy of emotions (ACC)}: We use emotion labels to calculate the accuracy of the emotions of generated sentences to reflect the controllability of emotion.

\begin{table*}[ht]
\resizebox{\linewidth}{!}{ 
\begin{tabular}{cp{\textwidth}} 
\toprule
\textbf{Context}    & Jessica had to go to the city.                                \\ \midrule \midrule
\textbf{SoCP}       & She was very excited to see a new . She was very proud of her friends . She was very happy . She was happy she was going to get .                               \\ 
\textbf{GPT-2}      & She left her friends and their cars behind. They waited outside the station for her. She knew she'd go to a bar one day. Unfortunately they all stayed away. \\ 
\textbf{BART}       & She told her mom to take a bus. The bus didn't have enough time to get back to the stations. When she came back her mother was upset. Jessica was upset that everyone didn't believe her to go now.               \\
\textbf{SSG}        &  She was leaving the bus at ten o'clock. She saw a traffic light coming up in the distance. But she got in the way quickly. She was late late to her bus stop so she had to wait.        \\ 
\hline 
\multirow{4}{*}{\textbf{Chae}}       & 1. $\langle SEP \rangle$ \textbf{$\langle soc \rangle$} Jessica \textbf{$\langle soa \rangle$} to learn new things $\langle sep \rangle$ to see the \textbf{museums} $\langle sep \rangle$ to learn something \textbf{$\langle soe \rangle$} \textbf{joy}  \\  
& 2. $\langle SEP \rangle$ \textbf{$\langle soc \rangle$} Jessica \textbf{$\langle soa \rangle$} to go out to some interesting place \textbf{$\langle soe \rangle$} joy \\
& 3. $\langle SEP \rangle$ \textbf{$\langle soc \rangle$} Jessica \textbf{$\langle soa \rangle$} to save the \textbf{artwork} $\langle sep \rangle$ to \textbf{remember} what she had seen \textbf{$\langle soe \rangle$} joy \\
& 4. $\langle SEP \rangle$ \textbf{$\langle soc \rangle$} \textbf{Jessica} \textbf{$\langle soa \rangle$} to \textbf{photoshoot} $\langle sep \rangle$ \textbf{to see the artistic beauty} inside the famous museum \textbf{$\langle soe \rangle$} joy
\\ \hline
\textbf{CHAE} & She had always \textbf{enjoyed} going to a \textbf{museum}. However, she could not walk with better to the \textbf{museum}. She had to \textbf{remember} all her \textbf{paintings} from this one place. \textbf{Jessica} \textbf{noticed an amazing view} that her \textbf{camera} had of a gallery.                                      \\ 
\hline
\textbf{Gold} & She went to some museums. She really enjoyed the artwork at the MET. She took a lot of photographs. She was there for hours. \\
\bottomrule
\end{tabular}}
\caption{\label{case-study}
Case study of the generated stories by our model and the baselines. The $Chae$ represents the assigned control conditions including characters, actions and emotions.
}
\end{table*}

\begin{table*}[ht]
\resizebox{\linewidth}{!}{ 
\begin{tabular}{cp{\textwidth}} 
\toprule
\textbf{Context}    & A polite thief was making robberies in the small town.              \\ \midrule \midrule
\textbf{Chae1}       & $\langle SEP \rangle$ $\langle soc \rangle$ People $\langle soa \rangle$ $\langle no\_action \rangle$ $\langle soe \rangle$ fear $\langle SEP \rangle$ $\langle soc \rangle$ \colorbox{yellow!50}{\textbf{Man}} $\langle soa \rangle$ to catch the thief $\langle soe \rangle$ \colorbox{orange!30}{\textbf{anger}}  \\ 
\textbf{Result1} & One day, a \colorbox{yellow!50}{\textbf{man}} walked up to him and \colorbox{orange!30}{\textbf{asked him to stop}}.   \\ 
\hline
\textbf{Chae2}  & $\langle SEP \rangle$ $\langle soc \rangle$ People $\langle soa \rangle$ $\langle no\_action \rangle$ $\langle soe \rangle$ fear $\langle SEP \rangle$ $\langle soc \rangle$ \colorbox{yellow!50}{\textbf{Man}} $\langle soa \rangle$ $\langle no\_action \rangle$ $\langle soe \rangle$ \colorbox{orange!30}{\textbf{joy}}  \\
\textbf{Result2} &  The \colorbox{yellow!50}{\textbf{man}} who was supposed to stop him was a \colorbox{orange!30}{\textbf{nice man}}.  \\
\hline
\textbf{Chae3}  & $\langle SEP \rangle$ $\langle soc \rangle$ People $\langle soa \rangle$ $\langle no\_action \rangle$ $\langle soe \rangle$ fear $\langle SEP \rangle$ $\langle soc \rangle$ \colorbox{yellow!50}{\textbf{Tom}} $\langle soa \rangle$ \colorbox{SkyBlue!40}{\textbf{to catch the thief}} $\langle soe \rangle$ \textbf{anger}  \\
\textbf{Result3} &  \colorbox{yellow!50}{\textbf{Tom}} decided to investigate and \colorbox{SkyBlue!40}{\textbf{caught the thief}}.  \\
\hline
\textbf{Chae4}  & $\langle SEP \rangle$ $\langle soc \rangle$ \colorbox{yellow!50}{\textbf{People}} $\langle soa \rangle$ \colorbox{SkyBlue!40}{\textbf{call the police}} $\langle soe \rangle$ \colorbox{orange!30}{\textbf{fear}} $\langle SEP \rangle$ $\langle soc \rangle$ \colorbox{yellow!50}{\textbf{Tom}} $\langle soa \rangle$ \colorbox{SkyBlue!40}{\textbf{call the police}} $\langle soe \rangle$ \colorbox{orange!30}{\textbf{anger}}  \\
\textbf{Result4} &  \textbf{Tom} \colorbox{SkyBlue!40}{\textbf{called the police}} and \colorbox{yellow!50}{\textbf{they}} \colorbox{orange!30}{\textbf{told him}} to \colorbox{SkyBlue!40}{\textbf{call the police}}.  \\
\bottomrule
\end{tabular}}
\caption{\label{controllability}
Case study of controllability.
}
\end{table*}

~\\
\noindent
\textbf{Results}~~
Table~\ref{auto-metrics} shows the automatic evaluation results. All baseline models are trained on our dataset ROCStory. 
Our model achieves the lowest Perplexity, which reflects the high quality of the stories generated by CHAE. Besides, the BART-based models (BART and SSG) are better than the GPT-2 model. The traditional seq2seq model SoCP has worse performance.
The greater BLEU scores of CHAE and its varieties imply stories closer to the golden truth, proving that the introduction of fine-grained control, mainly by means of the special input $Chae$, is conducive to improving the quality of the generated stories.
As for Distinct, the pre-trained models show excellent performance compared with the traditional seq2seq model. Our model attains the best score, which demonstrates CHAE's ability in generating more diverse stories.
We further compare the ACC with SoCP, and our model gets higher emotion accuracy.

Additionally, we also explore some decoding strategies including greedy search, beam search, top-k, and temperature, as shown in Table
\ref{Decoding-strategies}. The BLEU scores reach the best when we use the beam size equal to 3, while we get the best Distinct scores with the top-k equal to 3 and the temperature equal to 1.2.  The BLEU scores decrease with the increase of beam size. The BLEU scores with top-k are lower than beam search, but the Distinct scores are higher. When top-k is fixed, the higher the temperature, the lower the BLEU and the higher the Distinct. When the temperature is fixed, the higher the top-k, the lower the BLEU and the higher the Distinct.

\subsection{Ablation Studies}
As shown in Table \ref{auto-metrics}, we also conduct the ablation studies to verify the effectiveness of the additional control methods in our model. 
When we remove the copy mechanism \textbf{(w/o copy)}, the BLEU decreases and the Diversity increases, which suggests that the copy mechanism is beneficial in controlling content, but it also affects the diversity at the same time. In addition, we also observed that the copy mechanism can help the model to improve the accuracy of emotions.
On the contrary, when we remove the emotion loss \textbf{(w/o emo)}, the BLEU increases and the Diversity decreases, which reflects that the emotion loss does improve the diversity of the stories, but sacrifices controllability.
Furthermore, we remove both the copy mechanism and the emotion loss \textbf{(w/o copy w/o emo)}, which means just introducing the conditions $Chae$ to the vanilla BART. The results are still better than BART, illustrating the benefit brought by the fine-grained control conditions. 

In general, the copy mechanism and the emotion loss complement each other. The results show that the integrated model (CHAE) can obtain good scores and relatively balance on BLEU and Diversity, and has the best performance on PPL and ACC.

\subsection{Human Evaluation}
We conduct a human evaluation to compare CHAE with baselines on the following three metrics. (1) \textbf{Fluency}: The fluency of a sentence can reflect the quality of intra-sentence. (2) \textbf{Coherence}: The coherence of the story can reflect the cohesion of context and the quality of inter-sentence. (3) \textbf{Informativeness}: The good performance of a story in terms of informativeness indicates that there are a variety of rich words in the story. 
We recruit six annotators and divided them into two groups to annotate 50 stories randomly sampled. Each story is annotated by three workers to ensure fairness. The workers have two tasks: one is to compare the results generated by CHAE and other baselines, and the other is to score the controllability of CHAE according to whether the predicted sentences adhere to the assigned conditions $Chae$. 
The majority votes among the annotators will be the final decisions for the first task and we average the scores of all annotators for the second task.
The results are shown in Table \ref{Human}. Our model achieves the best performance in each metric compared with baselines and has 65\% controllability.


\subsection{Case Study}

\textbf{Comparison}~~
Table \ref{case-study} shows the stories generated from our model and the baselines. The story generated by the SoCP model reflects that the results generated by the models built on the seq2seq architecture are usually simple, short and less informative. The transformer-based models, such as GPT-2, BART, and SSG, show strong generative capabilities, which can generate coherent and informative stories. However, the content of these stories can not be controlled, and the characters' emotions and actions in the stories are not obvious. 

For our model, we can specify the characters, actions and emotions in the story that we desired, and the model can generate stories based on this information, i.e. $Chae$. The meaning of the four $Chae$ is to control the content of the next four sentences respectively.
The first sentence reflects that the character's action is to go to the “museum”, and the word “enjoyed” reflects the emotion of “joy”. The word “remember” and “artwork” in the third sentence corresponds to “remember” and “paintings” in $Chae3$. The model is connected to the “camera” from the “photoshoot” in $Chae4$, and the phrase “noticed an amazing view” corresponds to “to see the artistic beauty”, and the character “Jessica” appears in the sentence. The results show that our model can generate coherent and informative stories according to the control information.

~\\
\noindent 
\textbf{Controllability}~~
Table \ref{controllability} shows the examples of controllability. We use beam search (beam=2) to generate results instead of top-k, because the top-k method usually generates diverse words, which can not ensure controllability.
Our model generates sentences based on the same context and the given different $Chae$. 
The first two examples show that for the same character “Man”, given the emotion “anger” and “joy”, the model can generate sentences with corresponding emotions, such as the phrase “asked him to stop” and “he was a nice man”. In the third example, we change the character to “Tom”, and the generated sentence is also changed from “Man” to “Tom”, and the phrase “caught the thief” also reflects the action “to catch the thief”. In the last example we change Tom's action to “call the police”, and the resulting change from “caught the thief” to “called the police”. Also, we set people's actions as “call the police” and the emotion is fear, which results in the expression “they told him to call the police.”
The above reflects that our model has a good control effect on characters, actions and emotions.

\section{Conclusion and Future Work}

Through our model CHAE, we can create stories with fine-grained control according to the specified characters and corresponding actions and emotions, which is more convenient for practical applications, such as the creation of script novels, providing inspiration for screenwriters, and even acting as screenwriters in the future. 

However, our model also has some disadvantages: (1) The dataset contains only stories with 5 sentences, which is not enough for learning to generate longer stories. (2) The training of our model heavily relies on the annotations of characters, emotions, and actions in the dataset, while it is very expensive to obtain the annotated data. (3) The $Chae$ in the dataset actually has some noise, i.e., some descriptions in $Chae$ are not reflected in the corresponding sente
nces. (4) Our iterative generation method will result in a long time to train a story and may cause a cascade problem, which affects the overall quality of story generation.

In future work, we will adopt datasets with much more data and longer stories, such as \textbf{WritingPrompts} \citep{fan-etal-2018-hierarchical} and \textbf{WikiPlots}\footnote{\url{https://github.com/markriedl/WikiPlots}}. 
In addition, we will consider using commonsense reasoning techniques to reason about the emotions and actions of the characters in the story before further generating the story. 
Regarding the noise of $Chae$, we plan to conduct denoising in preprocessing to filter out the samples whose $Chae$ are inconsistent with the corresponding sentence. The problem can also be alleviated by dynamically controlling the weight of the conditions.
We are also further exploring more convenient and effective training methods to generate controllable stories by inputting control conditions in one go, rather than iterative generation.

\section*{Acknowledgements}
The work is partially supported by the National Nature Science Foundation of China (No. 61976160, 61906137, 61976158, 62076184, 62076182) and Shanghai Science and Technology Plan Project (No.21DZ1204800) and Technology research plan project of Ministry of Public and Security (Grant No.2020JSYJD01).

\bibliography{ref_story}
\bibliographystyle{acl_natbib}


\end{document}